\PassOptionsToPackage{table}{xcolor}
\documentclass[10pt,twocolumn,letterpaper]{article}

\usepackage{cvpr}      

\newcommand{\question}[2]{\noindent\textbf{\textit{#1}} \textit{#2}}

\usepackage{array}
\newcommand{\tablestyle}[2]{\setlength{\tabcolsep}{#1}\renewcommand{\arraystretch}{#2}\centering\footnotesize}
\newcolumntype{x}[1]{>{\centering\arraybackslash}p{#1pt}}
\newcolumntype{y}[1]{>{\raggedright\arraybackslash}p{#1pt}}
\newcolumntype{z}[1]{>{\raggedleft\arraybackslash}p{#1pt}}

\newlength\savewidth\newcommand\shline{\noalign{\global\savewidth\arrayrulewidth
  \global\arrayrulewidth 1pt}\hline\noalign{\global\arrayrulewidth\savewidth}}

\definecolor{cvprblue}{rgb}{0.21,0.49,0.74}
\usepackage[pagebackref,breaklinks,colorlinks,allcolors=cvprblue]{hyperref}

\usepackage{tabularx, multirow, stfloats}
\newcolumntype{C}{>{\centering \arraybackslash}X}

\def\paperID{16468} 
\def\confName{CVPR}
\def\confYear{2025}
\title{Exploring the Deep Fusion of Large Language Models and Diffusion Transformers for Text-to-Image Synthesis}

\author{
Bingda Tang~$^{1}$ \quad
Boyang Zheng~$^{1}$ \quad
Xichen Pan~$^{1}$ \quad
Sayak Paul~$^{2}$ \quad
Saining Xie~$^{1}$
\\
[2mm]
$^1$~New York University \quad
$^2$~Hugging Face
\\
}

\begin{document}
\maketitle
\begin{abstract}
This paper does not describe a new method; instead, it provides a thorough exploration of an important yet understudied design space related to recent advances in text-to-image synthesis---specifically, the deep fusion of large language models (LLMs) and diffusion transformers (DiTs) for multi-modal generation. Previous studies mainly focused on overall system performance rather than detailed comparisons with alternative methods, and key design details and training recipes were often left undisclosed. These gaps create uncertainty about the real potential of this approach. To fill these gaps, we conduct an empirical study on text-to-image generation, performing controlled comparisons with established baselines, analyzing important design choices, and providing a clear, reproducible recipe for training at scale. We hope this work offers meaningful data points and practical guidelines for future research in multi-modal generation. Code is available
at this repository: \url{https://github.com/tang-bd/fuse-dit}.
\end{abstract}
\section{Introduction}
\label{sec:introduction}

Text-to-image diffusion models have made remarkable progress in generating high-quality images from descriptive texts. Current state-of-the-art systems~\cite{flux, sd3, pixartsigma, sdxl, dalle3} typically derive text representations from specialized encoders, such as CLIP~\cite{clip} and T5~\cite{t5}. With the rise of decoder-only large language models (LLMs), there has been a growing amount of interest in their potential as replacements for these traditional text encoders~\cite{ella, sana1.5, lidit, luminanext, pgv3, llm4gen, dreamllm, kosmosg}. However, simply substituting LLMs has not yielded expected performance gains unless coupled with sophisticated architectural adaptations~\cite{lidit, sana, ella}. Prior work~\cite{lidit} attributes this to the misalignment between the next token prediction training objective of LLMs and the need for discriminative text representations in diffusion models. 

\begin{figure}
    \centering
    \includegraphics[width=\linewidth]{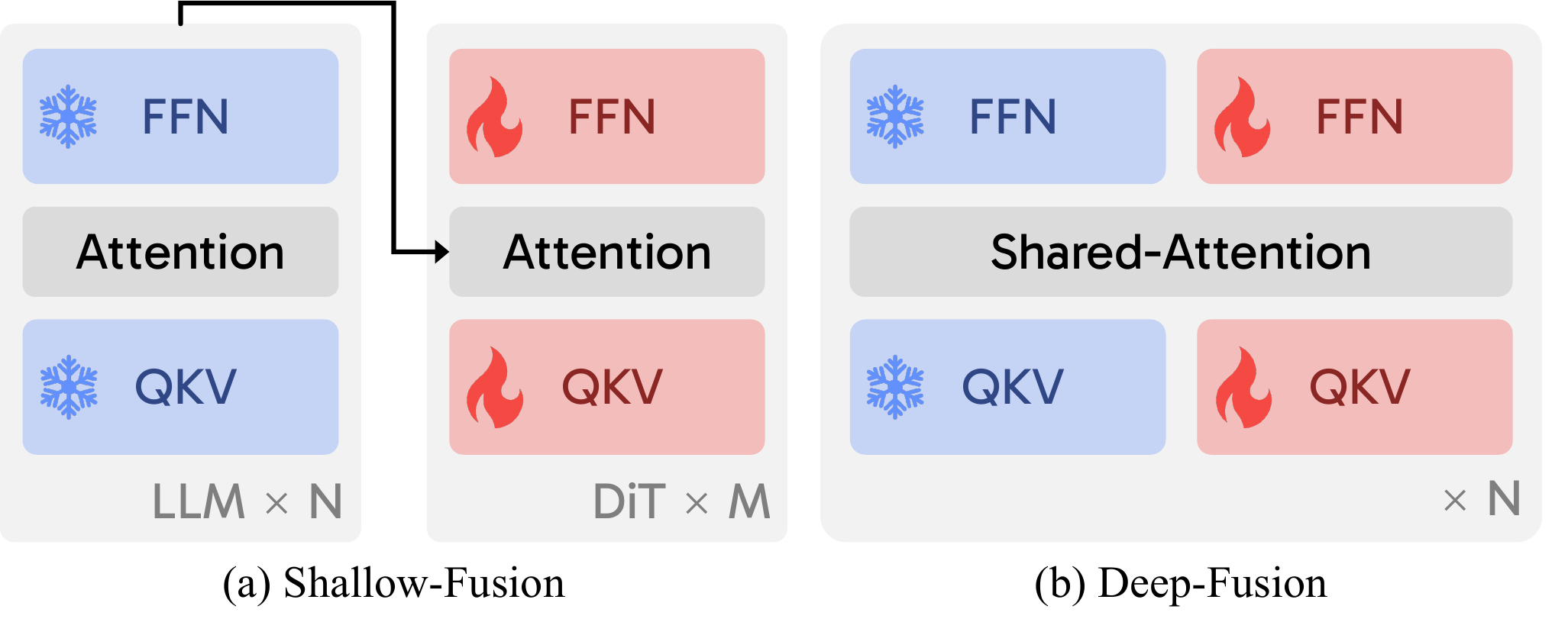}
    \vspace{-1.5em}
    \caption{
        \textbf{Illustration of the deep fusion approach and baselines.} We conduct controlled comparisons with baseline methods that incorporate text representations from a single text encoder layer into each DiT layer using late fusion within the attention mechanism, a strategy we term as the ``shallow fusion" approach.
    }
    \label{fig:deep_fusion}

    \vspace{-1.5em}
\end{figure}

Recent advancements~\cite{transfusion, omnigen, mot, janusflow} have successfully unified auto-regressive decoding and denoising diffusion within a single transformer~\cite{transformer}, enabling seamless multi-modal generation. This unified approach supports a wide range of tasks, including instructed image-to-text, text-to-image synthesis, and interleaved image-text generation. While earlier methods often relied on large-scale pretraining of the entire model, latest research~\cite{lmfusion, pgv3} flexibly leverages the computationally intensive pre-training of LLMs by deeply fusing them with diffusion transformers (DiTs)~\cite{dit} through layer-wise shared self-attention. This design facilitates rich cross-modal interactions while maintaining modality-specific computation by using distinct sets of weights. When optimized for text-to-image generation, it claims state-of-the-art performance~\cite{pgv3}.

Deep fusion presents a compelling alternative to existing architectures for text-to-image synthesis, which typically conditions directly on representations from a single text encoder layer. By aligning diffusion models with the auto-regressive decoding nature of LLMs, deep fusion enables a more natural and tight-knit use of these models. However, despite the existing positive signals, its true potential remains uncertain. Current research~\cite{lmfusion, pgv3} prioritizes system-level benchmarks over controlled comparisons with established baselines, obscuring its position within the broader research landscape. More critically, the design space remains severely underexplored, and essential implementation details, such as training recipes, are often undisclosed. These limitations impede reproducibility and hinder broader adoption within the research community.

In this paper, we bridge these gaps through an empirical study on the deep fusion of a frozen LLM and a trainable DiT for text-to-image synthesis. We conduct controlled comparisons between the deep fusion approach and baselines, examine key design choices, and introduce a scalable, reproducible training recipe that delivers competitive performance on the established benchmarks for text-to-image generation. We believe that the evidence and unresolved questions highlighted in this study are of significant importance. We hope this work serves as a valuable resource, offering meaningful data points and practical guidelines to drive future advancements in multi-modal generation. 
\section{Related Work}
\label{sec:related_work}
\paragraph{Conditioning mechanisms in diffusion models.}

Numerous studies have examined effective methods for integrating linguistic conditional information into diffusion models to facilitate text-to-image synthesis. Latent diffusion models (LDM)~\cite{ldm} pioneered the use of cross-attention between image and text features, a technique that has since become standard in U-Net~\cite{unet} based architectures~\cite{sdxl, pgv2.5}.

With the rise of DiTs~\cite{dit}, vision transformer~\cite{vit} has emerged as the dominant architecture for diffusion models, prompting a reassessment of conditioning mechanisms. The vanilla DiT employs adaLN-Zero modulation to inject conditional information. However, this approach is limited to using pooled text representations, which capture only coarse-grained information. Subsequent architectures have explored cross-attention~\cite{pixartalpha, pixartsigma, sana} and self-attention~\cite{lumina, luminanext, sd3, flux} for text conditioning, which typically extract representations from a single text encoder layer, usually the last or penultimate one. While this strategy aligns well with CLIP~\cite{clip} and T5~\cite{t5} text encoders, it is inherently mismatched with the next-token prediction training objective of LLMs, where the last layer focuses on next-token prediction instead of learning discriminative representation. In contrast, the deep fusion approach shows promise in harnessing the internal information flow of LLMs, aligning with their in-context self-attention mechanism for processing information.

\paragraph{Taming LLMs for diffusion models.}

The prevalence of decoder-only LLMs has driven extensive efforts to tame them for text-to-image diffusion models. The most effective and widely adopted practice involves leveraging LLMs to enrich input prompts~\cite{dalle3}, as well as employing multi-modal LLMs (MLLMs) to generate synthetic captions for image data~\cite{dalle3, sd3, pixartsigma, pgv3, sana}. Another popular direction integrates (M)LLMs as system components, such as planners and discriminators~\cite{rpg, llmgrounded, divideandconquer, selfcorrecting}. While these approaches have demonstrated effectiveness, they do not tap into the architecture of diffusion models.

To integrate LLMs into diffusion models, previous works attempt to replace text encoders with LLMs, either by training from scratch~\cite{sana, lidit, luminanext} or by aligning feature spaces~\cite{ella, dreamllm, kosmosg, llm4gen}. However, this substitution alone has not yielded the expected performance gains unless paired with sophisticated architectural adaptations~\cite{sana, ella, lidit}.

Recent research~\cite{transfusion, mot, omnigen, janusflow} has proven that auto-regressive decoding and denoising diffusion can be effectively unified within a single transformer for multi-modal generation. While earlier methods relied on large-scale pretraining of the entire model, latest research~\cite{lmfusion, pgv3} flexibly leverages the computationally intensive pretraining of LLMs by deeply fusing them with DiTs. From a text-to-image perspective, this architecture introduces a novel way to tame LLMs for text-to-image diffusion models, with the potential for achieving state-of-the-art performance~\cite{pgv3}.
\section{Deep Fusion of LLMs and DiTs}
\label{sec:approach}

\subsection{Model Architecture}

\begin{figure}
    \centering
    \includegraphics[width=0.6\linewidth]{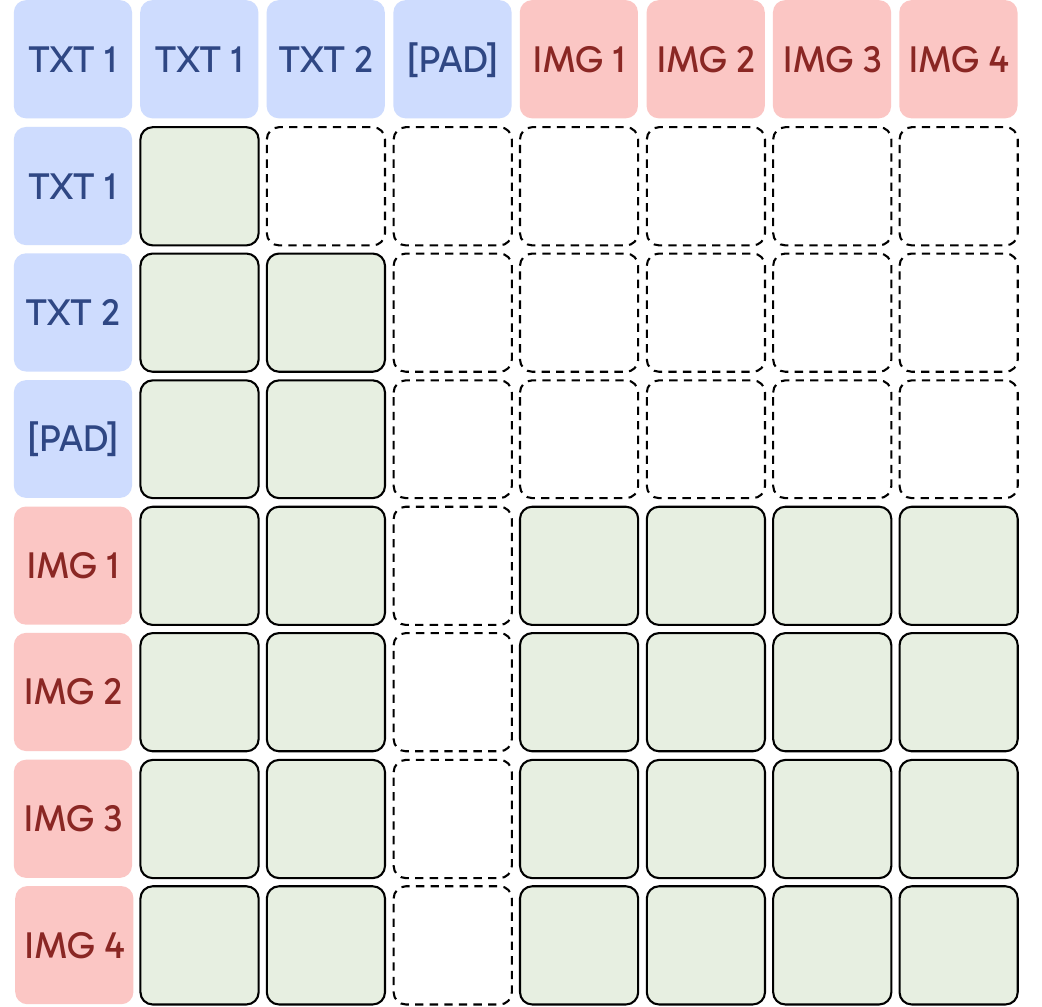}
    \caption{
        \textbf{Illustration of the attention mask.} Each dotted square indicates whether the row can attend to the column. 
    }
    \label{fig:attn-mask}
    \vspace{-1.5em}
\end{figure}

In the deep fusion approach, we integrate a frozen decoder-only LLM with a trainable DiT using layer-wise shared self-attention (\cref{fig:deep_fusion}). The DiT mirrors the LLM's transformer architecture, differing only in its input/output layers and timestep conditioning modules. This implements a two-stream transformer architecture that facilitates rich cross-modal interactions while maintaining modality-specific computation by utilizing distinct weight sets for processing tokens from different modalities.


The fused model processes text embeddings through the LLM stream and noisy image latents through the DiT stream. At each layer's self-attention operation, we concatenate token sequences from both streams, enabling the DiT to extract conditional information from the linguistic context. To preserve the pretrained LLM's functionality, we apply a causal attention mask to the text sequence and a bidirectional mask to the image sequence, permitting the image tokens to attend to text tokens but not vice versa (\cref{fig:attn-mask}). After the final layer, we discard text tokens and use only image tokens to predict velocity, as typically done when training rectified flow models~\cite{rectifiedflow}.

Notably, only the key and value states of the text hidden states are needed for the image tokens. These remain constant throughout the diffusion process, allowing them to be efficiently cached and reused during inference. 

\subsection{Training Objective}

We adopt the rectified flow formulation~\cite{rectifiedflow} to learn transport maps between the standard Gaussian noise distribution $\pi_0$ and the data distribution $\pi_1$, by connecting straight paths $x_t = tx_1 + (1-t)x_0$ between samples $x_0 \sim \pi_0, x_1 \sim \pi_1$ and learning an ODE model $dz_t = v_\theta(z_t, t)dt$ on time $t \in [0, 1]$ which converts $z_0$ from $\pi_0$ to a $z_1$ following $\pi_1$.

We fit the velocity $v$ with $x_1 - x_0$ under a prescribed time distribution $\pi_t$ by solving the following regression problem:
\vspace{-.5em}
\begin{equation}
\min_{v}\mathbb{E}_{t \sim \pi_t, x_0 \sim \pi_0, x_1 \sim \pi_1}[\Vert v(x_t, t) - (x_1 - x_0) \Vert^2]dt
\label{eq:rf}
\end{equation}

We parameterize $v$ with network $\theta$ and solve~\cref{eq:rf} by stochastic optimization with empirical draws. 
\section{Experiment Setup}
To ensure fully open and reproducible comparison between the deep and shallow fusion approaches, we provide comprehensive details on the experimental setup, including the model, dataset, training, inference, and evaluation. For the same purpose, we exclusively use open-source pre-trained LLMs and publicly available datasets.

\label{sec:experiments_1_setup}
\paragraph{Model.}
We employ a frozen Gemma 2B~\cite{gemma} as the base LLM for all experiments (excluding~\cref{sec:experiments_3}). We pair it with a randomly initialized 2.5B-parameter DiT. The transformer configurations of the DiT strictly follows the base LLM, including the hidden size, number of layers, number of attention heads, FFN design, and other architectural details, ensuring both models have an identical 2B-parameter backbone.
Following the vanilla DiT setup~\cite{dit}, we use 2D frequency absolute positional encoding, adaLN-Zero timestep-conditioning, ViT~\cite{vit}-style weight initialization, and a patch size of 2. 
To further stabilize training, we apply QK normalization to all layers.
For all experiments we adopt the same 16-channel VAE from Stable Diffusion 3 (SD 3)~\cite{sd3}. 

\paragraph{Dataset.} We use the CC12M~\cite{cc12m} dataset with community-sourced synthetic captions~\cite{cc12m-recaptioned} as our training set for all experiments excluding~\cref{sec:experiments_3}. Our downloaded version of the dataset includes 10.9M image-caption pairs.
The images are resized and center-cropped to $512 \times 512$ and the texts are padded or truncated to 256 tokens.

\paragraph{Training.} 
We train all models with a batch size of 512 using AdamW~\cite{adamw} optimizer ($\beta_1=0.9, \beta_2=0.999$) in BF16 mixed precision. We use a constant learning rate of $1 \times 10^{-4}$, a weight decay of $1 \times 10^{-4}$, and gradient clipping with a threshold of 1.0. Exponential moving average of the weights are gathered by a decay factor of 0.99 every 100 steps. We employ the same logit-normal distribution as used in SD 3 for timestep sampling. During training, 10\% of the texts are randomly dropped to learn unconditional generation. Training is carried out using Google TPU v4-256 pods and FSDP implemented by PyTorch / XLA SPMD.

\paragraph{Inference.} We conduct inference using Euler discretization with 25 sampling steps, and a classifier-free guidance scale of 6 which we find to be near optimal for text-image alignment. We employ identical sampling steps and guidance scale across all experiments.

\paragraph{Evaluation.} We evaluate image-text alignment using GenEval~\cite{geneval} and DPG-Bench~\cite{ella} metrics, prioritizing GenEval for its robustness. While both benchmarks provide valuable insights, DPG-Bench exhibits certain limitations, such as rapid performance saturation and potential measurement errors~\cite{pgv3}. To ensure a comprehensive evaluation, we also provide visual quality measurements  using FID~\cite{fid} on MJHQ-30K~\cite{pgv2.5}. Notably, image-text alignment does not always correlate positively with visual quality, often presenting trade-offs. Our sampling and evaluation are primarily carried out using NVIDIA L40S GPUs.

\section{Comparing Deep and Shallow Fusion}
\label{sec:experiments_1}
The deep fusion approach fuses an LLM and a DiT through layer-wise shared self-attention, creating interconnections throughout the network. However, established architectures typically condition on representations from a single text encoder layer. To investigate the true potential of deep fusion, we conduct controlled comparisons with baseline methods.

For a fair and meaningful comparison, we examine a common architectural paradigm which we refer to as shallow fusion. In this approach, representations from a single text encoder layer are integrated into each DiT layer through late fusion within the attention operation. Unlike deep fusion, which involves multi-layered interactions, shallow fusion maintains a fixed connection between each DiT layer and a prescribed text encoder layer (\cref{fig:deep_fusion}).

\subsection{Shallow Fusion Baselines}
\label{sec:experiments_1_baselines} 
\begin{figure}
    \centering
    \includegraphics[width=\linewidth]{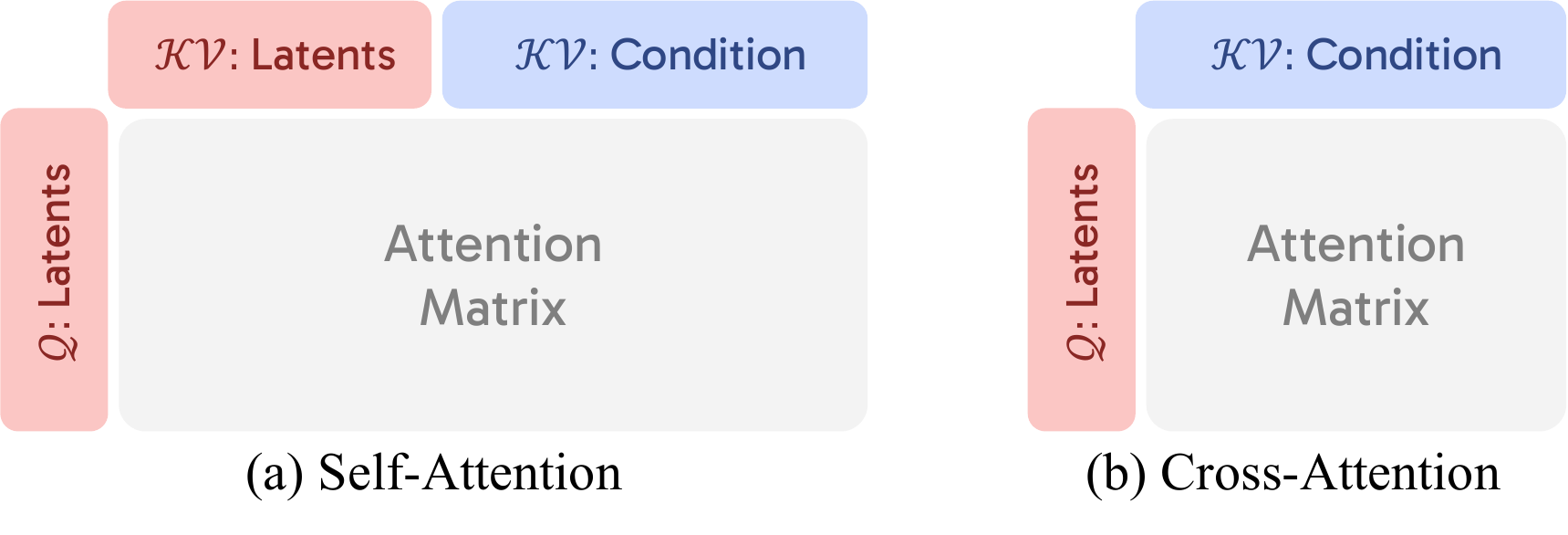}
    \vspace{-1.5em}
    \caption{
        \textbf{Illustration of cross-modal attention in the shallow fusion baselines.} The key and query states of the condition are directly projected from text representations.
    }
    \label{fig:attention_fuse}
    \vspace{-1.5em}
\end{figure}

We consider two shallow fusion architectures that condition on last-layer hidden states of LLMs as our baselines. As illustrated in \cref{fig:attention_fuse}, the two architecture differ in how they aggregate information from the condition.

\begin{itemize}
\item \textbf{Self-attention DiT.} 
In this design, text representations are projected to key and value states and then concatenated with those of image hidden states in self-attention, which can also be decoupled by running self-attention and cross-attention in parallel and merging their outputs. This approach resembles architectures proposed by ~\cite{lumina, luminanext}.

\item \textbf{Cross-attention DiT.}
This design also projects text representations to key and values states. However, unlike the previous approach, they are used for additional cross-attention with image hidden states, applied after the self-attention in each layer. This architecture follows the methodology employed in ~\cite{sana, pixartalpha, pixartsigma}.

\end{itemize}

Compared to the deep fusion approach, both baseline models include RMS normalization and a linear layer for text representations before passing them through additional key and value projection layers. Additionally, the cross-attention DiT model uses extra query projections in its cross-attention mechanism for image hidden states. Other model configurations follows the DiT design in~\cref{sec:experiments_1_setup}.

Notably, the deep fusion approach can be reinterpreted as a variant of the self-attention DiT architecture, as they both aggregate conditional information through in-context self-attention. The key difference lies in how the key and value states of the condition are derived: self-attention DiT employs a trainable projection to generate these states from a single text encoder layer, whereas deep fusion extracts them from corresponding LLM layers. Despite their similarities, the two approaches have fundamentally different conceptual implications: deep fusion treats the LLM and DiT as equal components of a unified model.

Apart from the aforementioned designs, SD3~\cite{sd3} introduced an alternative fusion strategy that uses a two-stream transformer (MM-DiT) to jointly processes noised image latents and linguistic representations. While this is another popular approach worthy of investigation, an apples-to-apples fair comparison between the deep fusion approach and MM-DiT is not feasible, as both streams in MM-DiT are trainable. Consequently, our analysis focuses solely on the shallow fusion baselines detailed in this section.

\subsection{Controlled Comparison}
\label{sec:experiments_1_comparison}
For a fair comparison, we design self-attention DiT, cross-attention DiT, and the deep fusion model with similar architectures, and train them for 300K steps following the setup detailed in~\cref{sec:experiments_1_setup}.

As shown in~\cref{tab:perf-comparison}, the deep fusion model achieve significantly better performance in image-text alignment than the self-attention DiT model and also surpass the cross-attention DiT model, while shallow fusion models demonstrate better visual quality. In terms of inference efficiency, deep fusion also demonstrates competitive performance, as shown in~\cref{tab:compute-comparison}. This positive evidence underscores the compelling positioning of the deep fusion approach within the current landscape.

\begin{table}[!htp]\vspace{-.7em}
\tablestyle{8pt}{1.05}
\begin{tabular}{y{60}|c|ccc}
Method & Params. &GenEval $\uparrow$ &DPG $\uparrow$ &FID $\downarrow$  \\
\shline
Self-Attention & 2.47B  & 0.42& 73.9 & 26.16\\
Cross-Attention & 2.62B & 0.49& 76.3 & \textbf{24.00} \\
Deep Fusion & 2.45B & \textbf{0.51}& \textbf{76.6} & 27.33\\
\end{tabular}
\vspace{-1em}
\caption{\textbf{Comparison of performance between deep and shallow fusion models.} Deep fusion beats shallow fusion in text-image alignment while underperforms in visual quality.}\label{tab:perf-comparison}
\vspace{-5pt}
\end{table}

\begin{table}[!htp]
\tablestyle{8pt}{1.05}\vspace{-1.5em}
\begin{tabular}{y{60}|c|c}
Method & Params. & Inference Latency (s)\\
\shline
Self-Attention & 2.47B & 1.75 \\
Cross-Attention & 2.62B & 1.86 \\
Deep Fusion & 2.45B & \textbf{1.66} \\
\end{tabular}
\vspace{-1em}
\caption{\textbf{Comparison of inference latency between deep and shallow fusion models.} The numbers are measured with a batch size of 1 in automatic mixed precision on an NVIDIA A100 GPU.}
    \label{tab:compute-comparison}
    \vspace{-2em}
\end{table}

\section{Examining Key Design Choices}
\label{sec:experiments_2}

In this section, we examine key design choices of the deep fusion approach through a text-to-image-centric lens. 
We begin by assessing the necessity and potential redundancy of parameters for timestep conditioning, determining whether to optimize their use or eliminate them altogether (\cref{sec:experiments_timestep}). Next, we compare various positional encoding strategies (\cref{sec:experiments_positional}). Additionally, we investigate how the choice of base LLM and the use of instruction prompts impacts text-to-image performance (\cref{sec:experiments_base}). 

For all experiments, we use the default design as the baseline and train the models for 300K steps, following the setup detailed in~\cref{sec:experiments_1_setup}.

\begin{figure}
    \centering
    \includegraphics[width=0.965\linewidth]{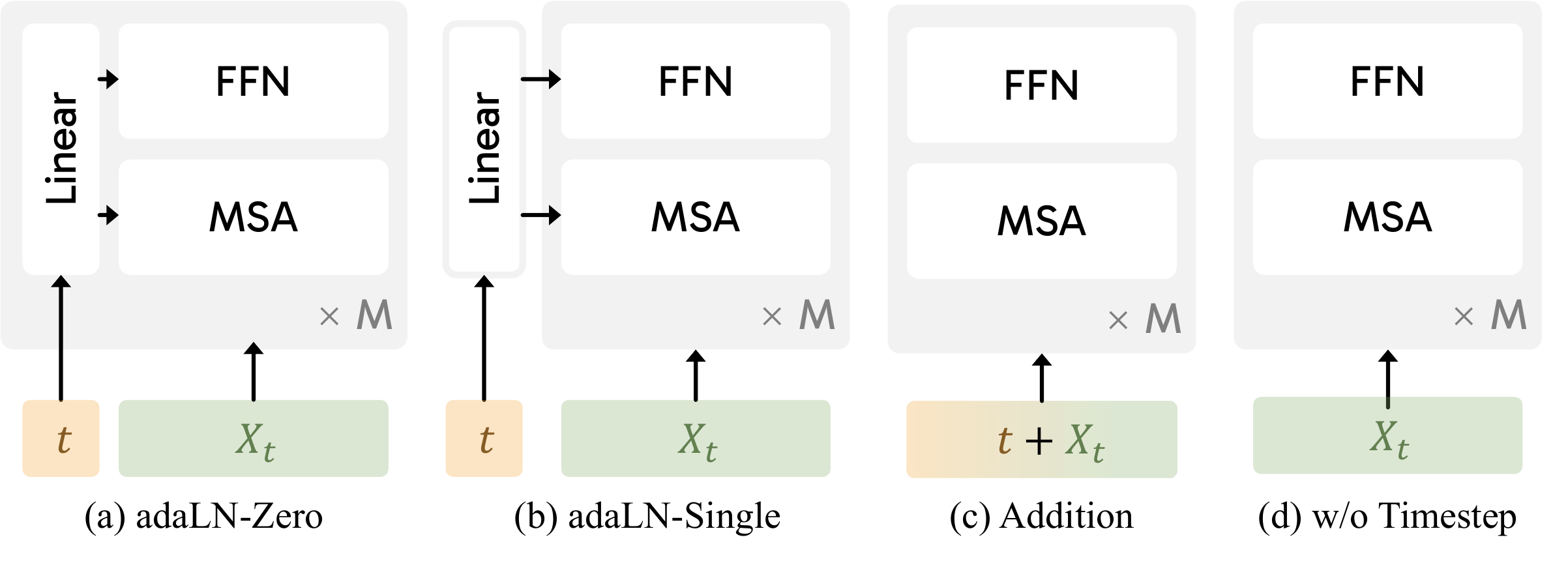}
    \vspace{-4pt}
    \caption{
        \textbf{Illustration of timestep conditioning strategies.} Removing timesetp conditioning leads to the fewest parameters and the best overall performance.
    }
    \label{fig:timestep_cond}
    \vspace{-1.5em}
\end{figure}

\subsection{Timestep Conditioning}
\label{sec:experiments_timestep}

DiT~\cite{dit} has introduced AdaLN-Zero as the standard mechanism for injecting timestep and class label information. The AdaLN modules typically accounts for a large proportion of the model parameters, 0.5B out of a total 2.5B in our case. However, since our text-to-image model does not use class labels, AdaLN serves only for timestep conditioning. This substantial parameter allocation raises a critical question about its necessity and potential redundancy: Could these parameters be utilized more effectively, or are they redundant altogether?

\question{Question 1.1.}{Can adaLN parameters be utilized more effectively by integrating additional text conditioning?}

Prior research \cite{flux, sd3, lumina} has integrated text information into AdaLN-Zero by augmenting timestep embeddings with pooled text representations through summation. To optimize parameter efficiency, we follow this approach by leveraging embeddings\footnote{We apply RMS-normalization and an MLP to the text embeddings before adding them to the timestep embeddings.} from the CLIP L/14 text encoder~\cite{clip}, which provides high-quality linguistic representations trained on large-scale multi-modal data. 

\begin{table}[!htp]\vspace{-.5em}
\centering
\tablestyle{8pt}{1.05}
\begin{tabular}{y{60}|ccc}
Method & GenEval $\uparrow$ & DPG $\uparrow$ & FID $\downarrow$ \\
\shline
\rowcolor{gray!30} adaLN-Zero & \textbf{0.51} & \textbf{76.6} & 27.33 \\
+ CLIP L/14 & 0.50 & 76.2 & \textbf{24.00} \\

\end{tabular}\vspace{-.5em}
\end{table}

The results indicate that integrating text modulation results in slight improvements in FID but weakens image-text alignment. Additionally, it further increases compute.

\question{Question 1.2.}{Can we shrink the parameters for timestep conditioning?}

We explore the possibility of eliminating timestep conditioning parameters to develop a more streamlined architecture akin to that of the LLM. Specifically, we compare four timestep conditioning strategies from previous work, each with a progressively reduced number of parameters, as illustrated in~\cref{fig:timestep_cond}.

\begin{itemize}
    \item \textbf{adaLN-Zero.} Following the vanilla DiT~\cite{dit}, we regress zero-initialized modulation parameters for each layer from the timestep embedding.
    \item \textbf{adaLN-Single.} Following PixArt-$\alpha$~\cite{pixartalpha}, we compute a global set of modulation parameters and refine them per layer by adding learnable embeddings.
    \item \textbf{Addition.} Following Transfusion~\cite{transfusion}, we directly add the timestep embedding to all image tokens.
    \item \textbf{w/o Timestep.} Inspired by a recent study~\cite{noise}, we completely remove the timestep conditioning from the model.
\end{itemize}

\begin{table}[!htp]
\tablestyle{8pt}{1.05}\vspace{-.3em}
\begin{tabular}{y{60}|c|ccc}
Method & Params. & GenEval $\uparrow$ & DPG $\uparrow$ & FID $\downarrow$ \\
\shline
\rowcolor{gray!30} adaLN-Zero & 2.47B & \textbf{0.51} & 76.6 & 27.33 \\
adaLN-Single & 2.01B & 0.47 & 75.2 & 27.09 \\
addition & 1.99B & 0.47 & 75.6 & 26.40 \\
w/o timestep & 1.98B & 0.49 & \textbf{76.7} & \textbf{21.27} \\

\end{tabular}

\vspace{-1em}
\caption{\textbf{Evaluation of timestep conditioning strategies.} Removing timestep conditioning yields surprisingly strong results.}\label{tab:timestep-conditioning}
\vspace{-.5em}
\end{table}

Surprisingly, the results in~\cref{tab:timestep-conditioning} indicate that reducing the number of parameters in timestep conditioning consistently enhances visual quality, whereas the performance in image-text alignment exhibits fluctuations. 

Notably, the strategy that completely removes timestep conditioning not only achieves significantly better FID but also maintains comparable GenEval and DPG-Bench performance. This finding is consistent with~\cite{noise}, where timestep conditioning removal improved FID in rectified flow models trained on smaller datasets. Furthermore, the complete removal of timestep conditioning eliminates the need for associated parameters, resulting in a 20\% reduction in the total number of model parameters. Although this approach slightly lags behind AdaLN-Zero in terms of text-image alignment metrics, we prefer it due to its parameter efficiency and architectural simplicity.

\begin{figure}
    \centering
    \includegraphics[width=\linewidth]{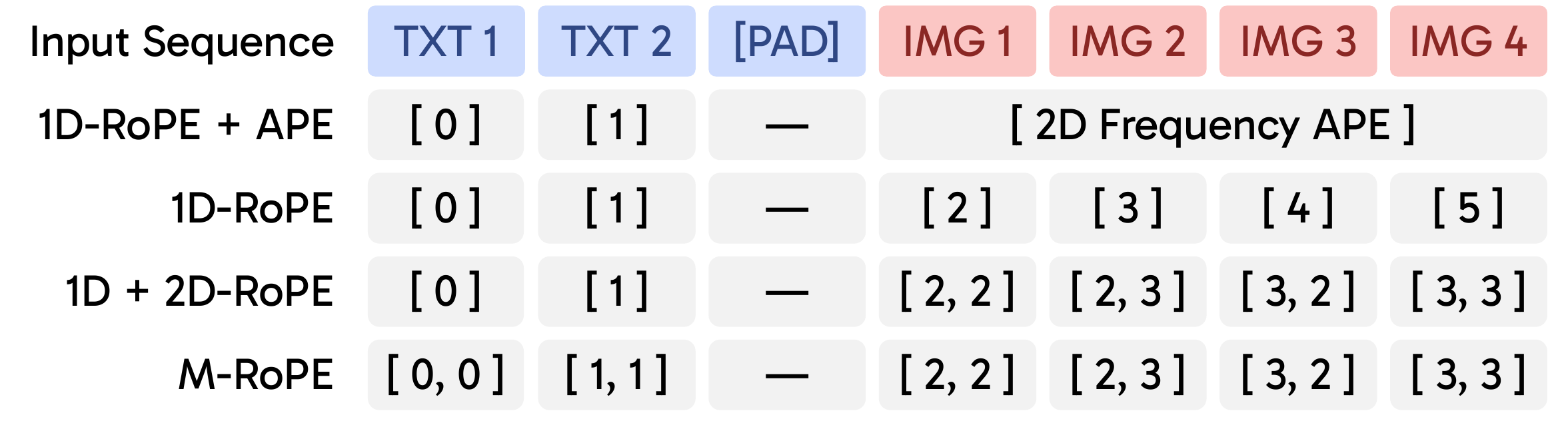}
    \vspace{-2em}
    \caption{
        \textbf{Illustration of RoPE.} The indices denote position IDs.
    }

    \label{fig:rope}
    \vspace{-1.5em}
\end{figure}

\subsection{Positional Encoding}
\label{sec:experiments_positional}
Absolute positional encoding (APE) is widely used in text-to-image diffusion models, whereas in the context of LLMs, rotary positional embedding (RoPE)~\cite{rope} is the predominant choice. 
As deep fusion models are inherently multi-modal and differ from traditional text-to-image diffusion models, it is unclear which positional encoding (or their combinations) is best suited for mixed-modal sequences. 

\question{Question 2.1.}{Is RoPE more advantageous than APE for enhancing the performance of deep fusion models?}

\begin{itemize}
    \item \textbf{1D RoPE + APE}: We apply 1D RoPE to the text sequence and APE to the image sequence respectively.
    \item \textbf{1D RoPE}: We extend 1D RoPE to encompass both text and image sequences.
    \item \textbf{1D + 2D RoPE}: We apply 1D RoPE to the text sequence and 2D RoPE to the image sequence respectively.
\end{itemize}

\begin{table}[!htp]\vspace{-.5em}
\tablestyle{8pt}{1.05}
\begin{tabular}{y{60}|ccc}
Method & GenEval $\uparrow$ & DPG $\uparrow$ & FID $\downarrow$ \\
\shline
\rowcolor{gray!30} 1D-RoPE + APE & \textbf{0.51} & 76.6 & 27.33 \\
1D-RoPE & 0.46 & \textbf{77.0} & 27.94 \\
1D + 2D-RoPE & \textbf{0.51} & 76.4 & \textbf{25.42} \\

\end{tabular}
\vspace{-1em}
\caption{\textbf{Comparing different positional encoding strategies.} 1D + 2D-RoPE achieves the best overall performance.}\label{tab:positional-encoding}
\end{table}\vspace{-.5em}

As shown in~\cref{tab:positional-encoding}, the 1D + 2D-RoPE configuration achieves the best overall performance, with only a marginal decrease in DPGBench compared to the 1D-RoPE + APE variant. The superiority of 2D-RoPE over APE suggests RoPE is more effective for modeling image sequences in deep fusion models. Using only 1D-RoPE slightly reduces performance, indicating that while deep fusion models treat text and image sequences as a unified input, their distinct positional characteristics are best modeled separately.

\question{Question 2.2.}{Do deep fusion models benefit from RoPE specifically designed for mixed-modal sequences?}

Previous work on MLLMs has explored RoPE strategies for handling mixed-modal sequences. Naturally, we are curious whether deep fusion models can benefit from these positional encodings. Follow Qwen2-VL, we implement M-RoPE, a variant of RoPE that applies 2D positional IDs to chunked 1D RoPE frequencies, allowing it to function as 1D RoPE for text sequences while approximating 2D RoPE for image sequences. 

\begin{center}\vspace{-0.2em}
\tablestyle{8pt}{1.05}
\begin{tabular}[t]{y{60}|ccc}
Method & GenEval $\uparrow$ & DPG $\uparrow$ & FID $\downarrow$ \\
\shline
\rowcolor{gray!30} 1D + 2D-RoPE & \textbf{0.51} & \textbf{76.4} & \textbf{25.42} \\
M-RoPE & 0.49 & 74.9 & 27.60 \\

\end{tabular}
\vspace{-0.2em}
\end{center}

Although M-RoPE elegantly unifies 1D and 2D-RoPEs, it still falls short compared to their direct combination. This underscores the challenge of designing position encodings for mixed-modal sequences.
\subsection{Base LLM}
\label{sec:experiments_base}

LLMs trained with different paradigms and data demonstrate diverse capabilities and behaviors. In this section, we examine how the choice of base LLM impacts the performance of deep fusion models.

\question{Question 3.1}{Can instruction tuning, combined with instruction prompts, improve text-to-image performance?}

Instruction tuning enables LLMs to effectively follow complex instructions. We explore whether this process can also enhance their internal information flow, leading to more contextualized and discriminative representations for text-to-image synthesis. We compare Gemma 2B with Gemma 2B IT, its instruction-tuned variant. We also experiment with using it with a simple instruction prompt, \textit{``Imagine: "}, which we find sufficient for guiding the LLM to generate detailed and relevant expansions of the input. 

\begin{table}[!htp]\vspace{-.5em}
\centering
\tablestyle{8pt}{1.05}
\begin{tabular}{y{80}|ccc}
Method & GenEval $\uparrow$ & DPG $\uparrow$ & FID $\downarrow$ \\
\shline
\rowcolor{gray!30} Gemma 2B & \textbf{0.51} & \textbf{76.6} & 27.33 \\
+ instruction tuning & 0.49 & 75.4 & 27.04 \\
~~~~ + instruction prompt & 0.50 & 75.8 & \textbf{25.28} \\
\end{tabular}
\vspace{-1em}
\caption{\textbf{Evaluating the effect of instruction tuning.} Using instruction-tuned LLMs does not improve performance.}\label{tab:gemma-it}
\vspace{-1em}
\end{table}

As shown in~\cref{tab:gemma-it}, instruction tuning appears to have a slightly negative impact on performance. While the use of an instruction prompt mitigates this effect to some extent, consistent with findings in~\cite{sana, lidit}, it still falls short of the baseline. This result highlights a challenge in effectively leveraging the instruction-following capabilities of LLMs.

\question{Question 3.2}{Can multi-modal tuning improve text-to-image performance?}

Additionally, we are interested in the effect of multi-modal tuning. While multi-modal tuning differs significantly from our setup, its shift in data distribution may still potentially enhance adaptability to multi-modal tasks. We compare Gemma 2B with the base LLM of PaliGemma 3B PT~\cite{paligemma}, a multi-modal extension of Gemma 2B that undergoes additional pretraining on image-text data.

\begin{center}\vspace{-.5em}
\tablestyle{8pt}{1.05}
\begin{tabular}[t]{y{80}|ccc}
Method & GenEval $\uparrow$ & DPG $\uparrow$ & FID $\downarrow$ \\
\shline
\rowcolor{gray!30} Gemma 2B & 0.51 & \textbf{76.6} & 27.33 \\
+ multi-modal tuning & \textbf{0.52} & 76.2 & \textbf{26.30} \\

\end{tabular}
\vspace{-.5em}
\end{center}

which yields small improvements in performance. This observation indicates that multi-modal finetuning provides some benefit to the deep-fusion model.

\question{Question 3.3}{Do improved LLM capabilities translate to stronger text-to-image performance?}

Finally, as the base LLM become more proficient in understanding and generating text, they could potentially foster synergistic improvements in DiT performance. We compare Gemma 2B with Gemma 2 2B~\cite{gemma2}, the next generation of Gemma 2B which demonstrates a 6\% absolute performance improvement ($0.44\to 0.50$) on an average of 8 language-only benchmarks. Notably, Gemma 2 2B features a different transformer architecture than Gemma 2B, prompting us to adjust the DiT architecture accordingly. This modification increases the number of adaLN parameters by 0.3B. However, as demonstrated in~\cref{sec:experiments_timestep}, these parameters do not contribute to model performance.

\begin{center}\vspace{-.4em}
\tablestyle{8pt}{1.05}
\begin{tabular}[t]{y{60}|ccc}
Model & GenEval $\uparrow$ & DPG $\uparrow$ & FID $\downarrow$ \\
\shline
\rowcolor{gray!30} Gemma 2B & 0.51 & 76.6 & 27.33 \\
Gemma 2 2B & \textbf{0.54} & \textbf{79.1} & \textbf{23.94} \\
\end{tabular}
\vspace{-.4em}
\end{center}

Upgrading from Gemma 2B to Gemma 2 2B yields a drastic performance boost. This finding suggests that the DiT's performance in deep fusion models is strongly dependent on the capabilities of the underlying base LLM.
\section{Training at Scale}
\label{sec:experiments_3}

In this section, we present a final recipe for the deep fusion model, building on the original framework while incorporating key insights from previous exploration. We conduct large-scale training to benchmark our model against established systems, showcasing its scalability and competitive performance on the leaderboard.

\subsection{Final Recipe}

Building on the insights from~\cref{sec:experiments_2}, we introduce the following design modifications to our model:

\begin{itemize}
    \item Remove AdaLN-Zero modules.
    \item Replace 1D-RoPE + APE with 1D + 2D-RoPE.
    \item Replace Gemma 2B with Gemma 2 2B, adjusting the DiT configurations accordingly.
\end{itemize}

We train our model, named FuseDiT, for 800K steps on a mixed dataset comprising CC12M~\cite{cc12m}, SA-1B~\cite{sam}, and the training subset of JourneyDB~\cite{journeydb}, amounting to approximately 26M image-caption pairs. Notably, state-of-the-art text-to-image models typically rely on high-quality datasets of much larger scale to achieve superior performance. For CC12M and SA-1B, we utilize synthetic captions~\cite{cc12m-recaptioned, pixartalpha}. Other experimental setup follows~\cref{sec:experiments_1_setup}.

\subsection{Performance Comparison}

\begin{table}[!htp]\vspace{-0.7em}
\centering
\fontsize{9pt}{9pt}\selectfont
\begin{tabularx}{\linewidth}{@{}lccccc@{}}
\toprule
Model & Params & Data &Gen.$\uparrow$ &DPG$\uparrow$ & FID$\downarrow$ \\
\midrule
SD 1.5~\cite{ldm} & 0.9B & 4.8B & 0.43 & 63.2& —  \\
DALL-E 2~\cite{dalle2} & 4.2B & 2.6B   & 0.52 & — & —\\
SDXL~\cite{sdxl} & 2.6B & 1.6B  &0.55 & 74.7& 6.63 \\
PG 2.5~\cite{pgv2.5} & 2.6B  & —  &0.56 & 75.5 & 6.09\\
SD 3 M~\cite{sd3} & 2B & 1B  &0.62 & 84.1 & 11.92\\
DALL-E 3~\cite{dalle3} & — & — &0.67 & 83.5 & —  \\
FLUX.1 [dev]~\cite{flux} & 12B  & — & 0.67 & 84.0 & 10.15 \\
PG 3~\cite{pgv3} & 24B & — &\textbf{0.76} & \textbf{87.0} & — \\
\midrule
MicroDiT~\cite{microdit} & 1.2B & 37M &0.46 & — & — \\
PixArt-$\alpha$~\cite{pixartalpha} & 0.6B & 25M&0.48 & 71.1 & — \\
Lumina Next~\cite{lumina} & 2B & — & 0.46 & 74.6& 7.58  \\
PixArt-$\Sigma$~\cite{pixartsigma} & 0.6B & 46M  & 0.54 & 80.5 & 6.15\\
Transfusion~\cite{transfusion} & 7.3B & 3.5B &0.63 & — & — \\
Sana 1.0 1.6B~\cite{sana} & 1.6B & — &0.66 & 84.8 & \textbf{5.76} \\
\midrule
\textbf{FuseDiT (Ours)} & 2B & 26M & 0.60 & 81.6 &7.54 \\
\bottomrule
\end{tabularx}
\caption{\textbf{Comparison of performance with state-of-the-art systems.} Table adapted from~\cite{sana1.5, transfusion}. Industrial baselines are presented at the top, while academic baselines and our model are listed at the bottom.}\label{tab:comparison}
\end{table}

We compare our model with the most advanced text-to-image diffusion models in~\cref{tab:comparison}. Despite being trained with limited compute and data in a simplified setting, our model surpasses many industry-standard systems and delivers competitive results.

\begin{figure*}
    \centering
    \includegraphics[width=\linewidth]{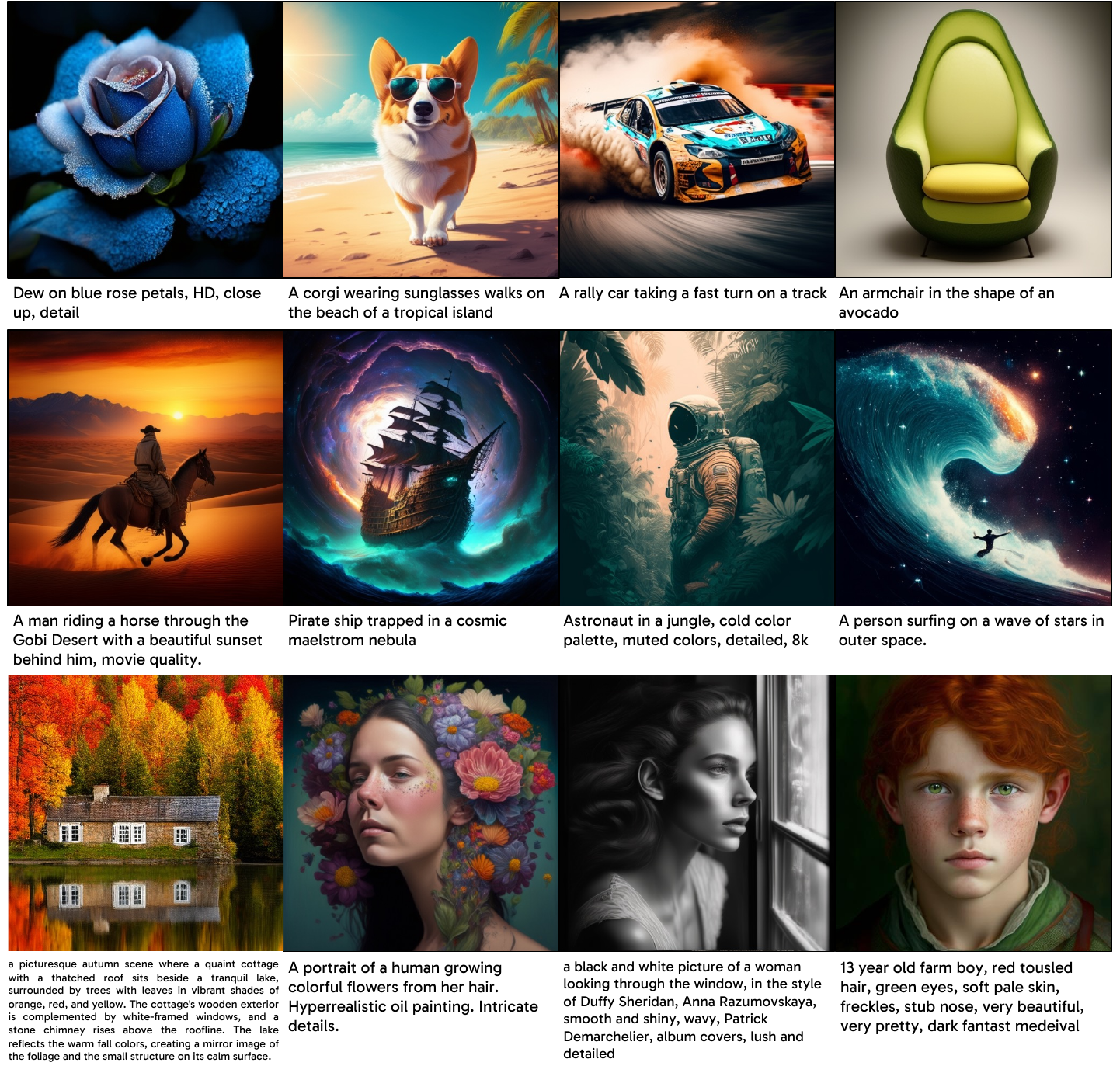}
    \caption{
        \textbf{Samples generated by FuseDiT.}
    }
    \label{fig:demo}
    \vspace{-1.5em}
\end{figure*}

We present qualitative examples from our model in~\Cref{fig:demo}. Our model demonstrates the ability to generate high-quality images with superior prompt alignment.

\section{Further Exploration}

In this section, we present preliminary studies exploring more aggressive modifications to the deep fusion approach. For all experiments, we train the models for 300K steps, following the setup detailed in~\cref{sec:experiments_1_setup}.

\begin{figure}
    \centering
    \includegraphics[width=\linewidth]{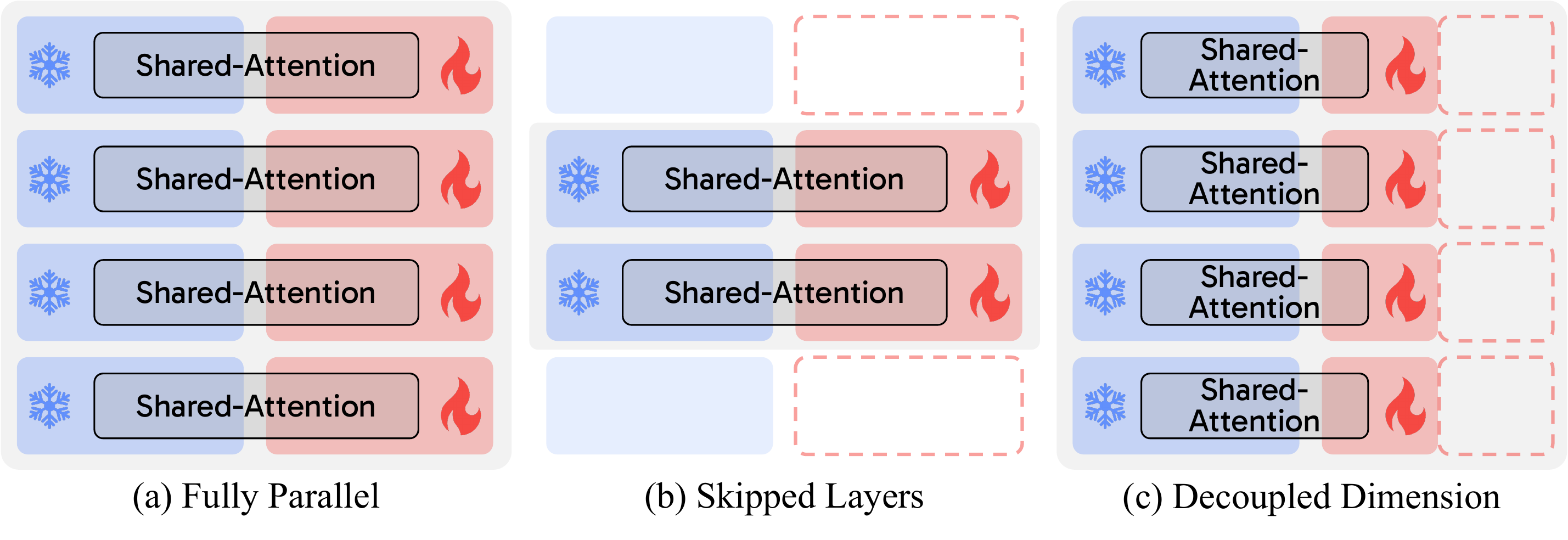}
    \caption{
        \textbf{Illustration of architecture alignment.} Dashed boxes indicate parameters that have been reduced by decreasing either the hidden size or the number of layers.
    }
    \label{fig:arch_align}
\vspace{-1.5em}
\end{figure}

\subsection{Architecture Alignment}
\label{sec:experiments_alignment}
Up to this point, our approach has followed prior work by aligning the LLM and DiT backbones in a layer-by-layer approach, strictly enforcing identical transformer configurations for both models. However, this rigid constraint limits the flexibility of deep fusion. In practice, we need the ability to scale the LLM and DiT independently, as different modalities follow distinct scaling laws and network design principles, and training and deployment scenarios vary. 

To address this, we explore modifying the DiT model’s hidden size and number of transformer layers, as illustrated in~\cref{fig:arch_align}. The adapted model is fused into the middle layers of the LLM, which contain richer semantic information~\cite{layerbylayer}. Additionally, in self-attention, hidden states are still projected to query, key, and value states that match the LLM’s dimensionality, ensuring compatibility. Since the most successful DiTs are generally much smaller than state-of-the-art LLMs, we focus on shrinking the size of our DiT.

\begin{table}[!htp]\vspace{-.5em}
\centering
\tablestyle{8pt}{1.05}
\begin{tabular}{y{40}|c|ccc}
Hidden size & Params. & GenEval $\uparrow$ & DPG $\uparrow$ & FID $\downarrow$ \\
\shline
\rowcolor{gray!30} 2048  & 2.5B & \textbf{0.51} & 76.6 & 27.33 \\
1792 & 2.1B & 0.50 & \textbf{77.1} & \textbf{24.27} \\
1536 & 1.8B & 0.49 & 76.2 & 25.46 \\
1280 & 1.4B &  0.48 & 74.8 & 24.64 \\

\end{tabular}
\vspace{-1em}
\caption{\textbf{Evaluating models of different hidden sizes.} The default hidden size is 2048.}\label{tab:hidden-size}
\end{table}

\begin{table}[!htp]\vspace{-1.7em}
\centering
\tablestyle{8pt}{1.05}
\begin{tabular}{y{40}|c|ccc}
Layers & Params. & GenEval $\uparrow$ & DPG $\uparrow$ & FID $\downarrow$ \\
\shline
\rowcolor{gray!30} 18  & 2.5B & \textbf{0.51} & \textbf{76.6} & 27.33 \\
14 & 1.9B & 0.47 & 74.6 & \textbf{23.46} \\
10 & 1.4B & 0.33 & 68.0 & 28.34 \\

\end{tabular}
\vspace{-1em}
\caption{\textbf{Evaluating different numbers of layers.} The default number of layer is 18.}\label{tab:num-layers}
\vspace{-.7em}
\end{table}

As shown in Table~\cref{tab:hidden-size}, the model's performance degrades gracefully as we reduce the hidden size, with visual quality actually improving in some cases. While decreasing the number of transformer layers (\cref{tab:num-layers}) also yields acceptable results, performance deteriorates more quickly. We hypothesize this occurs because Gemma 2B already employs fewer layers than typical model architectures of the same size. These findings suggest that LLM and DiT model designs can be effectively decoupled, enabling the application of separate scaling laws and design principles.

\subsection{Attention Mechanism}

In~\cref{sec:approach}, we built on prior work by defining the deep fusion architecture with shared self-attention to bridge the LLM and DiT. Inspired by extensive research on cross-attention in MLLMs~\cite{llama3, flamingo} and our findings in~\cref{sec:experiments_1}, which highlight the superior performance of cross-attention DiT over self-attention DiT, we explore an alternative deep fusion variant. This new approach replaces shared self-attention with cross-attention mechanisms, similar to cross-attention DiT but with a key distinction: we substitute the projected linguistic key and value states in traditional cross-attention DiT with corresponding states from LLM layers.
\begin{center}\vspace{5pt}
\tablestyle{8pt}{1.05}
\begin{tabular}{y{60}|ccc}
Method & GenEval $\uparrow$ & DPG $\uparrow$ & FID $\downarrow$ \\
\shline
self-attention & 0.51 & \textbf{76.6} & 27.33 \\
cross-attention & \textbf{0.52} & 76.5 & \textbf{26.57} \\

\end{tabular}
\end{center}

This modification yields minor gains, though at a cost to the LLM-DiT parity. Additionally, we find that although cross-attention introduces a negligible increase in FLOPs and parameter count, it leads to approximately a 12\% increase in latency\footnote{Latency is measured with a batch size of 1 under automatic mixed precision on an NVIDIA A100 GPU} (1.66s vs. 1.86s). Therefore, we retained the self-attention design in our final configuration.

\section{Conclusion}
\label{sec:conclusion}

We have studied the recently popular deep fusion of a frozen LLM with a trainable DiT for text-to-image synthesis. Our findings provide empirical evidence supporting its advantages over baselines. We highlight key design choices, identify unresolved problems, and offer meaningful data points alongside practical guidelines. We hope our empirical work help advance multi-modal generation and bridge the gap between auto-regressive decoding and denoising diffusion.

\clearpage

\section*{Acknowledgments}
We thank the reviewers for their constructive feedback. We also thank Shusheng Yang, Shengbang Tong, Wenhao Chai, Nanye Ma, Sihan Xu, and Chenyu Li for insightful discussions. This work was mainly supported by the Google TPU Research Cloud (TRC) program, the Google Cloud Research Credits program (GCP19980904), Open Path AI Foundation, and Lambda Labs. SX also acknowledges support from Intel AI SRS, IITP grant funded by the Korean Government (MSIT) (RS-2024-00457882, National AI Research Lab Project), Amazon Research Award, and NSF Award IIS-2443404. 
{
    \small
    \bibliographystyle{ieeenat_fullname}
    \bibliography{main}

\begin{thebibliography}{51}
\providecommand{\natexlab}[1]{#1}
\providecommand{\url}[1]{\texttt{#1}}
\expandafter\ifx\csname urlstyle\endcsname\relax
  \providecommand{\doi}[1]{doi: #1}\else
  \providecommand{\doi}{doi: \begingroup \urlstyle{rm}\Url}\fi

\bibitem[Alayrac et~al.(2022)Alayrac, Donahue, Luc, Miech, Barr, Hasson, Lenc, Mensch, Millican, Reynolds, et~al.]{flamingo}
Jean-Baptiste Alayrac, Jeff Donahue, Pauline Luc, Antoine Miech, Iain Barr, Yana Hasson, Karel Lenc, Arthur Mensch, Katherine Millican, Malcolm Reynolds, et~al.
\newblock Flamingo: a visual language model for few-shot learning.
\newblock In \emph{NeurIPS}, 2022.

\bibitem[Betker et~al.(2023)Betker, Goh, Jing, Brooks, Wang, Li, Ouyang, Zhuang, Lee, Guo, et~al.]{dalle3}
James Betker, Gabriel Goh, Li Jing, Tim Brooks, Jianfeng Wang, Linjie Li, Long Ouyang, Juntang Zhuang, Joyce Lee, Yufei Guo, et~al.
\newblock Improving image generation with better captions.
\newblock \emph{https://cdn. openai. com/papers/dall-e-3. pdf}, 2023.

\bibitem[Beyer et~al.(2024)Beyer, Steiner, Pinto, Kolesnikov, Wang, Salz, Neumann, Alabdulmohsin, Tschannen, Bugliarello, et~al.]{paligemma}
Lucas Beyer, Andreas Steiner, Andr{\'e}~Susano Pinto, Alexander Kolesnikov, Xiao Wang, Daniel Salz, Maxim Neumann, Ibrahim Alabdulmohsin, Michael Tschannen, Emanuele Bugliarello, et~al.
\newblock Paligemma: A versatile 3b vlm for transfer.
\newblock \emph{arXiv:2407.07726}, 2024.

\bibitem[Changpinyo et~al.(2021)Changpinyo, Sharma, Ding, and Soricut]{cc12m}
Soravit Changpinyo, Piyush Sharma, Nan Ding, and Radu Soricut.
\newblock {Conceptual 12M}: Pushing web-scale image-text pre-training to recognize long-tail visual concepts.
\newblock In \emph{CVPR}, 2021.

\bibitem[Chen et~al.(2024{\natexlab{a}})Chen, Ge, Xie, Wu, Yao, Ren, Wang, Luo, Lu, and Li]{pixartsigma}
Junsong Chen, Chongjian Ge, Enze Xie, Yue Wu, Lewei Yao, Xiaozhe Ren, Zhongdao Wang, Ping Luo, Huchuan Lu, and Zhenguo Li.
\newblock Pixart-$\sigma$: Weak-to-strong training of diffusion transformer for 4k text-to-image generation.
\newblock In \emph{ECCV}, 2024{\natexlab{a}}.

\bibitem[Chen et~al.(2024{\natexlab{b}})Chen, Yu, Ge, Yao, Xie, Wang, Kwok, Luo, Lu, and Li]{pixartalpha}
Junsong Chen, Jincheng Yu, Chongjian Ge, Lewei Yao, Enze Xie, Zhongdao Wang, James Kwok, Ping Luo, Huchuan Lu, and Zhenguo Li.
\newblock Pixart-$\alpha$: Fast training of diffusion transformer for photorealistic text-to-image synthesis.
\newblock In \emph{ICLR}, 2024{\natexlab{b}}.

\bibitem[Dong et~al.(2024)Dong, Han, Peng, Qi, Ge, Yang, Zhao, Sun, Zhou, Wei, et~al.]{dreamllm}
Runpei Dong, Chunrui Han, Yuang Peng, Zekun Qi, Zheng Ge, Jinrong Yang, Liang Zhao, Jianjian Sun, Hongyu Zhou, Haoran Wei, et~al.
\newblock Dreamllm: Synergistic multimodal comprehension and creation.
\newblock In \emph{ICLR}, 2024.

\bibitem[Dosovitskiy et~al.(2021)Dosovitskiy, Beyer, Kolesnikov, Weissenborn, Zhai, Unterthiner, Dehghani, Minderer, Heigold, Gelly, Uszkoreit, and Houlsby]{vit}
Alexey Dosovitskiy, Lucas Beyer, Alexander Kolesnikov, Dirk Weissenborn, Xiaohua Zhai, Thomas Unterthiner, Mostafa Dehghani, Matthias Minderer, Georg Heigold, Sylvain Gelly, Jakob Uszkoreit, and Neil Houlsby.
\newblock An image is worth 16x16 words: Transformers for image recognition at scale.
\newblock In \emph{ICLR}, 2021.

\bibitem[Emporium(2024)]{cc12m-recaptioned}
Caption Emporium.
\newblock conceptual-captions-cc12m-llavanext.
\newblock \url{https://huggingface.co/datasets/CaptionEmporium/conceptual-captions-cc12m-llavanext}, 2024.

\bibitem[Esser et~al.(2024)Esser, Kulal, Blattmann, Entezari, M{\"u}ller, Saini, Levi, Lorenz, Sauer, Boesel, et~al.]{sd3}
Patrick Esser, Sumith Kulal, Andreas Blattmann, Rahim Entezari, Jonas M{\"u}ller, Harry Saini, Yam Levi, Dominik Lorenz, Axel Sauer, Frederic Boesel, et~al.
\newblock Scaling rectified flow transformers for high-resolution image synthesis.
\newblock In \emph{ICML}, 2024.

\bibitem[Gao et~al.(2024)Gao, Zhuo, Lin, Liu, Chen, Du, Xie, Luo, Qiu, Zhang, et~al.]{lumina}
Peng Gao, Le Zhuo, Ziyi Lin, Chris Liu, Junsong Chen, Ruoyi Du, Enze Xie, Xu Luo, Longtian Qiu, Yuhang Zhang, et~al.
\newblock Lumina-t2x: Transforming text into any modality, resolution, and duration via flow-based large diffusion transformers.
\newblock \emph{arXiv:2405.05945}, 2024.

\bibitem[Ghosh et~al.(2023)Ghosh, Hajishirzi, and Schmidt]{geneval}
Dhruba Ghosh, Hanna Hajishirzi, and Ludwig Schmidt.
\newblock Geneval: An object-focused framework for evaluating text-to-image alignment.
\newblock \emph{arXiv:2310.11513}, 2023.

\bibitem[Grattafiori et~al.(2024)Grattafiori, Dubey, Jauhri, Pandey, Kadian, Al-Dahle, Letman, Mathur, Schelten, Vaughan, et~al.]{llama3}
Aaron Grattafiori, Abhimanyu Dubey, Abhinav Jauhri, Abhinav Pandey, Abhishek Kadian, Ahmad Al-Dahle, Aiesha Letman, Akhil Mathur, Alan Schelten, Alex Vaughan, et~al.
\newblock The llama 3 herd of models.
\newblock \emph{arXiv:2407.21783}, 2024.

\bibitem[Heusel et~al.(2017)Heusel, Ramsauer, Unterthiner, Nessler, and Hochreiter]{fid}
Martin Heusel, Hubert Ramsauer, Thomas Unterthiner, Bernhard Nessler, and Sepp Hochreiter.
\newblock Gans trained by a two time-scale update rule converge to a local nash equilibrium.
\newblock In \emph{NeurIPS}, 2017.

\bibitem[Hu et~al.(2024)Hu, Wang, Fang, Fu, Cheng, and Yu]{ella}
Xiwei Hu, Rui Wang, Yixiao Fang, Bin Fu, Pei Cheng, and Gang Yu.
\newblock Ella: Equip diffusion models with llm for enhanced semantic alignment.
\newblock \emph{arXiv:2403.05135}, 2024.

\bibitem[Kirillov et~al.(2023)Kirillov, Mintun, Ravi, Mao, Rolland, Gustafson, Xiao, Whitehead, Berg, Lo, et~al.]{sam}
Alexander Kirillov, Eric Mintun, Nikhila Ravi, Hanzi Mao, Chloe Rolland, Laura Gustafson, Tete Xiao, Spencer Whitehead, Alexander~C Berg, Wan-Yen Lo, et~al.
\newblock Segment anything.
\newblock In \emph{ICCV}, 2023.

\bibitem[Labs(2024)]{flux}
Black~Forest Labs.
\newblock Announcing black forest labs.
\newblock \url{https://blackforestlabs.ai/announcing-black-forest-labs/}, 2024.

\bibitem[Li et~al.(2024)Li, Kamko, Akhgari, Sabet, Xu, and Doshi]{pgv2.5}
Daiqing Li, Aleks Kamko, Ehsan Akhgari, Ali Sabet, Linmiao Xu, and Suhail Doshi.
\newblock Playground v2. 5: Three insights towards enhancing aesthetic quality in text-to-image generation.
\newblock \emph{arXiv:2402.17245}, 2024.

\bibitem[Lian et~al.(2023)Lian, Li, Yala, and Darrell]{llmgrounded}
Long Lian, Boyi Li, Adam Yala, and Trevor Darrell.
\newblock Llm-grounded diffusion: Enhancing prompt understanding of text-to-image diffusion models with large language models.
\newblock \emph{arXiv:2305.13655}, 2023.

\bibitem[Liang et~al.(2024)Liang, Yu, Luo, Iyer, Dong, Zhou, Ghosh, Lewis, Yih, Zettlemoyer, et~al.]{mot}
Weixin Liang, Lili Yu, Liang Luo, Srinivasan Iyer, Ning Dong, Chunting Zhou, Gargi Ghosh, Mike Lewis, Wen-tau Yih, Luke Zettlemoyer, et~al.
\newblock Mixture-of-transformers: A sparse and scalable architecture for multi-modal foundation models.
\newblock \emph{arXiv:2411.04996}, 2024.

\bibitem[Liu et~al.(2024{\natexlab{a}})Liu, Akhgari, Visheratin, Kamko, Xu, Shrirao, Souza, Doshi, and Li]{pgv3}
Bingchen Liu, Ehsan Akhgari, Alexander Visheratin, Aleks Kamko, Linmiao Xu, Shivam Shrirao, Joao Souza, Suhail Doshi, and Daiqing Li.
\newblock Playground v3: Improving text-to-image alignment with deep-fusion large language models.
\newblock \emph{arXiv:2409.10695}, 2024{\natexlab{a}}.

\bibitem[Liu et~al.(2024{\natexlab{b}})Liu, Ma, Zhang, Zhen, Zhao, Hu, Liu, and Fan]{llm4gen}
Mushui Liu, Yuhang Ma, Xinfeng Zhang, Yang Zhen, Zeng Zhao, Zhipeng Hu, Bai Liu, and Changjie Fan.
\newblock Llm4gen: Leveraging semantic representation of llms for text-to-image generation.
\newblock \emph{arXiv:2407.00737}, 2024{\natexlab{b}}.

\bibitem[Liu et~al.(2022)Liu, Gong, and Liu]{rectifiedflow}
Xingchao Liu, Chengyue Gong, and Qiang Liu.
\newblock Flow straight and fast: Learning to generate and transfer data with rectified flow.
\newblock \emph{arXiv:2209.03003}, 2022.

\bibitem[Loshchilov and Hutter(2019)]{adamw}
Ilya Loshchilov and Frank Hutter.
\newblock Decoupled weight decay regularization.
\newblock In \emph{ICLR}, 2019.

\bibitem[Ma et~al.(2024{\natexlab{a}})Ma, Zong, Song, Li, and Liu]{lidit}
Bingqi Ma, Zhuofan Zong, Guanglu Song, Hongsheng Li, and Yu Liu.
\newblock Exploring the role of large language models in prompt encoding for diffusion models.
\newblock \emph{arXiv:2406.11831}, 2024{\natexlab{a}}.

\bibitem[Ma et~al.(2024{\natexlab{b}})Ma, Liu, Chen, Liu, Wu, Wu, Pan, Xie, Zhang, Zhao, et~al.]{janusflow}
Yiyang Ma, Xingchao Liu, Xiaokang Chen, Wen Liu, Chengyue Wu, Zhiyu Wu, Zizheng Pan, Zhenda Xie, Haowei Zhang, Liang Zhao, et~al.
\newblock Janusflow: Harmonizing autoregression and rectified flow for unified multimodal understanding and generation.
\newblock \emph{arXiv:2411.07975}, 2024{\natexlab{b}}.

\bibitem[Mesnard et~al.(2024)Mesnard, Hardin, Dadashi, Bhupatiraju, Pathak, Sifre, Rivi{\`e}re, Kale, Love, et~al.]{gemma}
Thomas Mesnard, Cassidy Hardin, Robert Dadashi, Surya Bhupatiraju, Shreya Pathak, Laurent Sifre, Morgane Rivi{\`e}re, Mihir~Sanjay Kale, Juliette Love, et~al.
\newblock Gemma: Open models based on gemini research and technology.
\newblock \emph{arXiv:2403.08295}, 2024.

\bibitem[Pan et~al.(2024)Pan, Dong, Huang, Peng, Chen, and Wei]{kosmosg}
Xichen Pan, Li Dong, Shaohan Huang, Zhiliang Peng, Wenhu Chen, and Furu Wei.
\newblock Kosmos-g: Generating images in context with multimodal large language models.
\newblock In \emph{ICLR}, 2024.

\bibitem[Peebles and Xie(2023)]{dit}
William Peebles and Saining Xie.
\newblock Scalable diffusion models with transformers.
\newblock In \emph{ICCV}, 2023.

\bibitem[Podell et~al.(2023)Podell, English, Lacey, Blattmann, Dockhorn, M{\"u}ller, Penna, and Rombach]{sdxl}
Dustin Podell, Zion English, Kyle Lacey, Andreas Blattmann, Tim Dockhorn, Jonas M{\"u}ller, Joe Penna, and Robin Rombach.
\newblock Sdxl: Improving latent diffusion models for high-resolution image synthesis.
\newblock \emph{arXiv:2307.01952}, 2023.

\bibitem[Radford et~al.(2021)Radford, Kim, Hallacy, Ramesh, Goh, Agarwal, Sastry, Askell, Mishkin, Clark, Krueger, and Sutskever]{clip}
Alec Radford, Jong~Wook Kim, Chris Hallacy, Aditya Ramesh, Gabriel Goh, Sandhini Agarwal, Girish Sastry, Amanda Askell, Pamela Mishkin, Jack Clark, Gretchen Krueger, and Ilya Sutskever.
\newblock Learning transferable visual models from natural language supervision.
\newblock In \emph{ICML}, 2021.

\bibitem[Raffel et~al.(2020)Raffel, Shazeer, Roberts, Lee, Narang, Matena, Zhou, Li, and Liu]{t5}
Colin Raffel, Noam Shazeer, Adam Roberts, Katherine Lee, Sharan Narang, Michael Matena, Yanqi Zhou, Wei Li, and Peter~J Liu.
\newblock Exploring the limits of transfer learning with a unified text-to-text transformer.
\newblock In \emph{JMLR}, 2020.

\bibitem[Ramesh et~al.(2022)Ramesh, Dhariwal, Nichol, Chu, and Chen]{dalle2}
Aditya Ramesh, Prafulla Dhariwal, Alex Nichol, Casey Chu, and Mark Chen.
\newblock Hierarchical text-conditional image generation with clip latents.
\newblock \emph{arXiv:2204.06125}, 2022.

\bibitem[Riviere et~al.(2024)Riviere, Pathak, Sessa, Hardin, Bhupatiraju, Hussenot, Mesnard, Shahriari, Ram{\'e}, et~al.]{gemma2}
Morgane Riviere, Shreya Pathak, Pier~Giuseppe Sessa, Cassidy Hardin, Surya Bhupatiraju, L{\'e}onard Hussenot, Thomas Mesnard, Bobak Shahriari, Alexandre Ram{\'e}, et~al.
\newblock Gemma 2: Improving open language models at a practical size.
\newblock \emph{arXiv:2408.00118}, 2024.

\bibitem[Rombach et~al.(2022)Rombach, Blattmann, Lorenz, Esser, and Ommer]{ldm}
Robin Rombach, Andreas Blattmann, Dominik Lorenz, Patrick Esser, and Bj{\"o}rn Ommer.
\newblock High-resolution image synthesis with latent diffusion models.
\newblock In \emph{CVPR}, 2022.

\bibitem[Ronneberger et~al.(2015)Ronneberger, Fischer, and Brox]{unet}
Olaf Ronneberger, Philipp Fischer, and Thomas Brox.
\newblock U-net: Convolutional networks for biomedical image segmentation.
\newblock In \emph{MICCAI}, 2015.

\bibitem[Sehwag et~al.(2024)Sehwag, Kong, Li, Spranger, and Lyu]{microdit}
Vikash Sehwag, Xianghao Kong, Jingtao Li, Michael Spranger, and Lingjuan Lyu.
\newblock Stretching each dollar: Diffusion training from scratch on a micro-budget.
\newblock \emph{arXiv:2407.15811}, 2024.

\bibitem[Shi et~al.(2024)Shi, Han, Zhou, Liang, Lin, Zettlemoyer, and Yu]{lmfusion}
Weijia Shi, Xiaochuang Han, Chunting Zhou, Weixin Liang, Xi~Victoria Lin, Luke Zettlemoyer, and Lili Yu.
\newblock Lmfusion: Adapting pretrained language models for multimodal generation.
\newblock \emph{arXiv:2412.15188}, 2024.

\bibitem[Skean et~al.(2025)Skean, Arefin, Zhao, Patel, Naghiyev, LeCun, and Shwartz-Ziv]{layerbylayer}
Oscar Skean, Md~Rifat Arefin, Dan Zhao, Niket Patel, Jalal Naghiyev, Yann LeCun, and Ravid Shwartz-Ziv.
\newblock Layer by layer: Uncovering hidden representations in language models.
\newblock \emph{arXiv:2502.02013}, 2025.

\bibitem[Su et~al.(2024)Su, Ahmed, Lu, Pan, Bo, and Liu]{rope}
Jianlin Su, Murtadha Ahmed, Yu Lu, Shengfeng Pan, Wen Bo, and Yunfeng Liu.
\newblock Roformer: Enhanced transformer with rotary position embedding.
\newblock \emph{Neurocomputing}, 2024.

\bibitem[Sun et~al.(2024)Sun, Pan, Ge, Li, Duan, Wu, Zhang, Zhou, Qin, Wang, et~al.]{journeydb}
Keqiang Sun, Junting Pan, Yuying Ge, Hao Li, Haodong Duan, Xiaoshi Wu, Renrui Zhang, Aojun Zhou, Zipeng Qin, Yi Wang, et~al.
\newblock Journeydb: A benchmark for generative image understanding.
\newblock In \emph{NeurIPS}, 2024.

\bibitem[Sun et~al.(2025)Sun, Jiang, Zhao, and He]{noise}
Qiao Sun, Zhicheng Jiang, Hanhong Zhao, and Kaiming He.
\newblock Is noise conditioning necessary for denoising generative models?
\newblock \emph{arXiv:2502.13129}, 2025.

\bibitem[Vaswani et~al.(2017)Vaswani, Shazeer, Parmar, Uszkoreit, Jones, Gomez, Kaiser, and Polosukhin]{transformer}
Ashish Vaswani, Noam Shazeer, Niki Parmar, Jakob Uszkoreit, Llion Jones, Aidan~N. Gomez, Lukasz Kaiser, and Illia Polosukhin.
\newblock Attention is all you need.
\newblock In \emph{NeurIPS}, 2017.

\bibitem[Wang et~al.(2024)Wang, Xie, Li, Wang, Liu, and Li]{divideandconquer}
Zhenyu Wang, Enze Xie, Aoxue Li, Zhongdao Wang, Xihui Liu, and Zhenguo Li.
\newblock Divide and conquer: Language models can plan and self-correct for compositional text-to-image generation.
\newblock \emph{arXiv:2401.15688}, 2024.

\bibitem[Wu et~al.(2024)Wu, Lian, Gonzalez, Li, and Darrell]{selfcorrecting}
Tsung-Han Wu, Long Lian, Joseph~E Gonzalez, Boyi Li, and Trevor Darrell.
\newblock Self-correcting llm-controlled diffusion models.
\newblock In \emph{CVPR}, 2024.

\bibitem[Xiao et~al.(2024)Xiao, Wang, Zhou, Yuan, Xing, Yan, Li, Wang, Huang, and Liu]{omnigen}
Shitao Xiao, Yueze Wang, Junjie Zhou, Huaying Yuan, Xingrun Xing, Ruiran Yan, Chaofan Li, Shuting Wang, Tiejun Huang, and Zheng Liu.
\newblock Omnigen: Unified image generation.
\newblock \emph{arXiv:2409.11340}, 2024.

\bibitem[Xie et~al.(2024)Xie, Chen, Chen, Cai, Lin, Zhang, Li, Lu, and Han]{sana}
Enze Xie, Junsong Chen, Junyu Chen, Han Cai, Yujun Lin, Zhekai Zhang, Muyang Li, Yao Lu, and Song Han.
\newblock Sana: Efficient high-resolution image synthesis with linear diffusion transformers.
\newblock \emph{arXiv:2410.10629}, 2024.

\bibitem[Xie et~al.(2025)Xie, Chen, Zhao, Yu, Zhu, Lin, Zhang, Li, Chen, Cai, et~al.]{sana1.5}
Enze Xie, Junsong Chen, Yuyang Zhao, Jincheng Yu, Ligeng Zhu, Yujun Lin, Zhekai Zhang, Muyang Li, Junyu Chen, Han Cai, et~al.
\newblock Sana 1.5: Efficient scaling of training-time and inference-time compute in linear diffusion transformer.
\newblock \emph{arXiv:2501.18427}, 2025.

\bibitem[Yang et~al.(2024)Yang, Yu, Meng, Xu, Ermon, and Cui]{rpg}
Ling Yang, Zhaochen Yu, Chenlin Meng, Minkai Xu, Stefano Ermon, and Bin Cui.
\newblock Mastering text-to-image diffusion: Recaptioning, planning, and generating with multimodal llms.
\newblock In \emph{ICML}, 2024.

\bibitem[Zhou et~al.(2024)Zhou, Yu, Babu, Tirumala, Yasunaga, Shamis, Kahn, Ma, Zettlemoyer, and Levy]{transfusion}
Chunting Zhou, Lili Yu, Arun Babu, Kushal Tirumala, Michihiro Yasunaga, Leonid Shamis, Jacob Kahn, Xuezhe Ma, Luke Zettlemoyer, and Omer Levy.
\newblock Transfusion: Predict the next token and diffuse images with one multi-modal model.
\newblock \emph{arXiv:2408.11039}, 2024.

\bibitem[Zhuo et~al.(2024)Zhuo, Du, Xiao, Li, Liu, Huang, Liu, Zhao, Wang, Ma, et~al.]{luminanext}
Le Zhuo, Ruoyi Du, Han Xiao, Yangguang Li, Dongyang Liu, Rongjie Huang, Wenze Liu, Lirui Zhao, Fu-Yun Wang, Zhanyu Ma, et~al.
\newblock Lumina-next: Making lumina-t2x stronger and faster with next-dit.
\newblock \emph{arXiv:2406.18583}, 2024.

\end{thebibliography}
}

\end{document}